\documentclass{article}

\usepackage{arxiv}

\usepackage[utf8]{inputenc} 
\usepackage[T1]{fontenc}    
\usepackage{hyperref}       
\usepackage{url}            
\usepackage{booktabs}       
\usepackage{amsfonts}       
\usepackage{nicefrac}       
\usepackage{microtype}      
\usepackage{lipsum}		
\usepackage{amsthm}
\usepackage{graphicx}
\usepackage{doi}
\usepackage{physics}
\usepackage{amssymb}
\usepackage{pifont}%

\newcommand{\pd}[2]{\frac{\partial #1}{\partial #2}}
\newcommand{\pdd}[2]{\frac{\partial^2 #1}{\partial #2^2}}

\usepackage{setspace}
\title{On the failure of ReLU activation for physics-informed machine learning}

\date{} 					

\author{
    Conor Rowan \\
	Smead Aerospace Engineering Sciences\\
	University of Colorado Boulder\\
    3775 Discovery Drive\\
	Boulder, CO 80309 \\
	\texttt{conor.rowan@colorado.edu} \\
}

\renewcommand{\headeright}{}
\renewcommand{\undertitle}{}
\renewcommand{\shorttitle}{}

\hypersetup{
pdftitle={A template for the arxiv style},
pdfsubject={q-bio.NC, q-bio.QM},
pdfauthor={David S.~Hippocampus, Elias D.~Striatum},
pdfkeywords={First keyword, Second keyword, More},
}

\begin{document}
\maketitle

\begin{abstract}
    Physics-informed machine learning uses governing ordinary and/or partial differential equations to train neural networks to represent the solution field. Like any machine learning problem, the choice of activation function influences the characteristics and performance of the solution obtained from physics-informed training. Several studies have compared common activation functions on benchmark differential equations, and have unanimously found that the rectified linear unit (ReLU) is outperformed by competitors such as the sigmoid, hyperbolic tangent, and swish activation functions. In this work, we diagnose the poor performance of ReLU on physics-informed machine learning problems. While it is well-known that the piecewise linear form of ReLU prevents it from being used on second-order differential equations, we show that ReLU fails even on variational problems involving only first derivatives. We identify the cause of this failure as second derivatives of the activation, which are taken not in the formulation of the loss, but in the process of training. Namely, we show that automatic differentiation in PyTorch fails to characterize derivatives of discontinuous fields, which causes the gradient of the physics-informed loss to be mis-specified, thus explaining the poor performance of ReLU.   
\end{abstract}



\section{Introduction}

\paragraph{} Hello reader. The purpose of this short paper is to explain a curious failure of the Deep Ritz method to recover the solution of a very simple boundary value problem when using ReLU activation functions. As always, the story begins with a one-dimensional elliptic boundary value problem:

\begin{equation}\label{strong_form}
    \pd{}{x}\qty(\kappa(x)\pd{u}{x}) + b(x) = 0, \quad x\in[0,1], \quad u(0)=u(1)=0.
\end{equation}

The physics here are beside the point---it suffices to say that $x$ is the spatial coordinate, $u(x)$ is the solution field, $b(x)$ is a source term, and $\kappa(x)$ is a coefficient field reminiscent of a material property. We call this the ``strong form'' of the governing equation. As is well-known in the engineering mechanics community \cite{bauchau_structural_2009}, obtaining a solution to Eq. \eqref{strong_form} can be equivalently thought of as the minimization of a particular ``energy'' functional:

\begin{equation}\label{energy}
    \Pi( v(x) ) = \int_0^1 \frac{1}{2} \kappa(x) \qty(\pd{v}{x})^2 - b(x) v(x) dx, \quad u(x) = \underset{v(x)}{\text{argmin }} \Pi( v(x) ).
\end{equation}

The calculus of variations can be used to show that the minimizer of the variational energy $\Pi$ satisfies an Euler-Lagrange equation given by Eq. \eqref{strong_form} \cite{goldstein_classical_2002}. We call Eq. \eqref{energy} the ``variational form'' of the problem. Regardless of the perspective we take on the solution, the general lack of analytical solutions to Eqs. \eqref{strong_form} and \eqref{energy} necessitates that we resort to approximations. Most traditional numerical methods represent the unknown solution field in a basis $\{ f_i(x)\}_{i=1}^N$, where $N$ is the number of basis functions and each $f_i$ is chosen by the analyst. The representation of the solution is thus $u(x) = \sum_{i=1}^N \theta_i f_i(x)$, where $\boldsymbol \theta$ are the unknown degrees of freedom which specify the approximation. By way of example, spectral methods \cite{ghanem_stochastic_1991} and finite element methods \cite{hughes_thomas_j_r_finite_2000} are two classical strategies which differ only in their choice of basis.

\paragraph{} Traditional numerical methods have been tremendously successful. We can simulate tiny volumes of gas for minuscule amounts of time \cite{lai_molecular_2018}, understand how bodies jiggle using linear viscoelasticity \cite{mavlanov_dynamic_2025}, and design lightweight aircraft against fracture\footnote{Though I am told that Boeing still relies on a complex variable ansatz solution to an infinite linear elastic plate from the 1930s \cite{westergaard_bearing_1939}.} \cite{bazant_critical_2022}.  Some go so far as to say that yesterday's science has been so successful that there is nothing fundamental left to discover \cite{horgan_end_2017}. Not so, says the practitioner of machine learning! In fact, a number of tools have been developed expressly for the purpose of facilitating the next generation of fundamental scientific discoveries \cite{brunton_discovering_2016, cranmer_interpretable_2023, udrescu_ai_2020}, though what precisely we are trying to discover remains unspecified. Furthermore, physics-informed neural networks (PINNs) were introduced between 2017 and 2019 to remedy the headaches and frustration that arose from the complex indexing procedures and laborious hand calculations germane to the finite element method \cite{raissi_physics-informed_2019, sirignano_dgm_2018}. The original PINN formulation represents the solution field with a multi-layer perceptron (MLP) neural network $\mathcal N(x;\boldsymbol \theta)$, where the parameter vector $\boldsymbol \theta$ is the collection of weights and biases. The MLP network consists of the repeated application of the following transformation:

\begin{equation*}
    \mathbf{y}_k = \sigma\Big(  \mathbf{A}_k\mathbf{x}_k + \mathbf{b}_k  \Big),
\end{equation*}

\noindent where $\mathbf{x}_k$ is the input to the $k$-th layer, $\mathbf{A}_k$ are the weights, and $\mathbf{b}_k$ are the biases. The output $\mathbf{y}_k$ becomes the input to the next layer $\mathbf{y}_k \rightarrow \mathbf{x}_{k+1}$. The function $\sigma(\cdot)$ is a nonlinear activation function that is applied element-wise. To solve differential equations with the standard PINNs approach, an optimization objective is formed by taking the strong form of Eq. \eqref{strong_form}, plugging in the neural network discretization, squaring the discretized differential equation pointwise, and integrating it over the domain. The parameters are then chosen such that this objective is minimized. Whereas boundary conditions were originally treated with additional penalties in the objective, an elegant alternative for the case of Dirichlet boundaries is given by \cite{sukumar_exact_2022}, where the boundary conditions are enforced automatically by multiplying the neural network with a ``distance function'' $D(x)$. For our one-dimensional problem with homogeneous Dirichlet boundaries, the solution is thus represented as

\begin{equation*}
    u(x) = \sin (\pi x) \mathcal N(x; \boldsymbol \theta),
\end{equation*}

\noindent where $\sin(\pi x)$ is the distance function. This same discretization can be used in the ``Deep Ritz'' formulation of the problem \cite{e_deep_2017}, which obtains a numerical solution through minimizing the variational energy of Eq. \eqref{energy} in terms of the neural network parameters. 

\paragraph{} Whether physics-informed or not, the choice of activation function plays a fundamental role in the training and performance of a neural network. Some common choices of activation functions are hyperbolic tangent ($\tanh(\bullet)$), sigmoid ($1/(1+\exp(-\bullet))$), softplus ($\ln ( 1 + \exp(\bullet))$), and the rectified linear unit (ReLU) ($\max(0,\bullet)$). Other options include the swish activation and the Gaussian error linear unit (GELU) \cite{hendrycks_gaussian_2023}. The problem of choosing activation functions has been studied extensively in the physics-informed machine learning literature. For example, in \cite{zafar_optimizing_2025}, using non-standard ``hybrid'' activation functions was shown to reduce errors in PINN solutions. Similarly, \cite{jagtap_adaptive_2020} showed that allowing the activation functions themselves to be partially learnable accelerated convergence in the physics-informed training. Another study found that learning the activation function allowed the network to better adapt to characteristics of the solution field \cite{wang_learning_2023}. For high-frequency solution fields, modifications inspired by the SIREN activation \cite{sitzmann_implicit_2020} were shown to reduce the number of parameters and thus expedite training \cite{fazliani_enhancing_2025}. In a comparative study on the Helmholtz equation, the swish activation function outperformed other activation functions by obtaining the smallest error with a reference solution \cite{al-safwan_is_2021}. In another work, the one-dimensional wave equation was used to compare the convergence and errors obtained from different activation functions, including hyperbolic tangent, swish, and sinusoidal activation functions, finding that no single activation performed best in all numerical experiments \cite{maczuga_influence_2023}. The authors remark that their choice to use the strong form of the governing equation places restrictions on the set of available activation functions. Namely, the governing equation requires second derivatives of the solution field, and the second derivative of ReLU is zero everywhere except at the origin. As such, they do not include ReLU in their study. Finally, \cite{dung_choice_nodate} compared six different activation functions including hyperbolic tangent, ReLU, swish, and modifications thereof on incompressible flow problems, finding that ReLU consistently performed the worst.

\paragraph{} None of the studies mentioned above conclude that ReLU is the optimal activation function for physics-informed training. Some authors indicate that higher-order continuity of the activation function---which ReLU does not possess---is a boon in the context of physics-informed training \cite{so_higher-order-relu-kans_2024}. To illustrate this, consider the strong form loss of Eq. \eqref{strong_form} and a neural network discretization using ReLU activation. We denote the ReLU activation function by $R(x) = \max(0,x)$. A single hidden-layer MLP network with ReLU is given by 

\begin{equation}\label{1hl}
    u(x) = \sum_{i=1}^N \theta^3_i R( \theta^1_i x + \theta^2_i),
\end{equation}

\noindent where $\boldsymbol \theta = [ \boldsymbol \theta^1 , \boldsymbol \theta^2 , \boldsymbol \theta^3]^T\in \mathbb R^{3N}$ is the collection of trainable parameters. Substituting Eq. \eqref{1hl} into the strong form of Eq. \eqref{strong_form} and using the chain rule, we see that the second derivative of the solution involves terms multiplying $\partial^2 R /\partial x^2$, which is zero everywhere except the origin. Because ReLU gives rise to piecewise-linear solution fields, it does not have a well-defined second derivative and is thus not suitable for use with the strong form. However, the variational energy of Eq. \eqref{energy} requires only first derivatives of the solution field. Relying on this formulation of the problem is one of the ways that piecewise linear hat functions can be employed in finite element methods. Because the first derivative of Eq. \eqref{1hl} does not vanish, and because piecewise linear representations of solution fields are used extensively in finite element methods, we expect the ReLU discretization to be well-suited for the variational objective. To investigate whether this is the case, we solve the variational optimization problem of Eq. \eqref{energy} with the following problem parameters:

\begin{equation}\label{1hldist}
    \kappa(x) = 1, \quad b(x) = 100 \sin(3\pi x), \quad u(x) = \sum_{i=1}^N \theta^3_i \sin(\pi x)R( \theta^1_i x + \theta^2_i) ,\quad N=50.
\end{equation}

This is a single hidden-layer ReLU network of width $50$. Notice that we introduce a distance function to enforce the homogeneous Dirichlet boundaries. This technically voids the argument that the second spatial derivative of the solution field is almost everywhere zero, however, we note that the curvature of the solution is controlled to a large extent by the distance function---an undesirable property of a discretization. If this argument is unsatisfying to the reader, we will see shortly that even the first-order energy formulation with the discretization of Eq. \eqref{1hldist} presents problems. 

\begin{figure}[hbt!]
\centering
\includegraphics[width=0.99\textwidth]{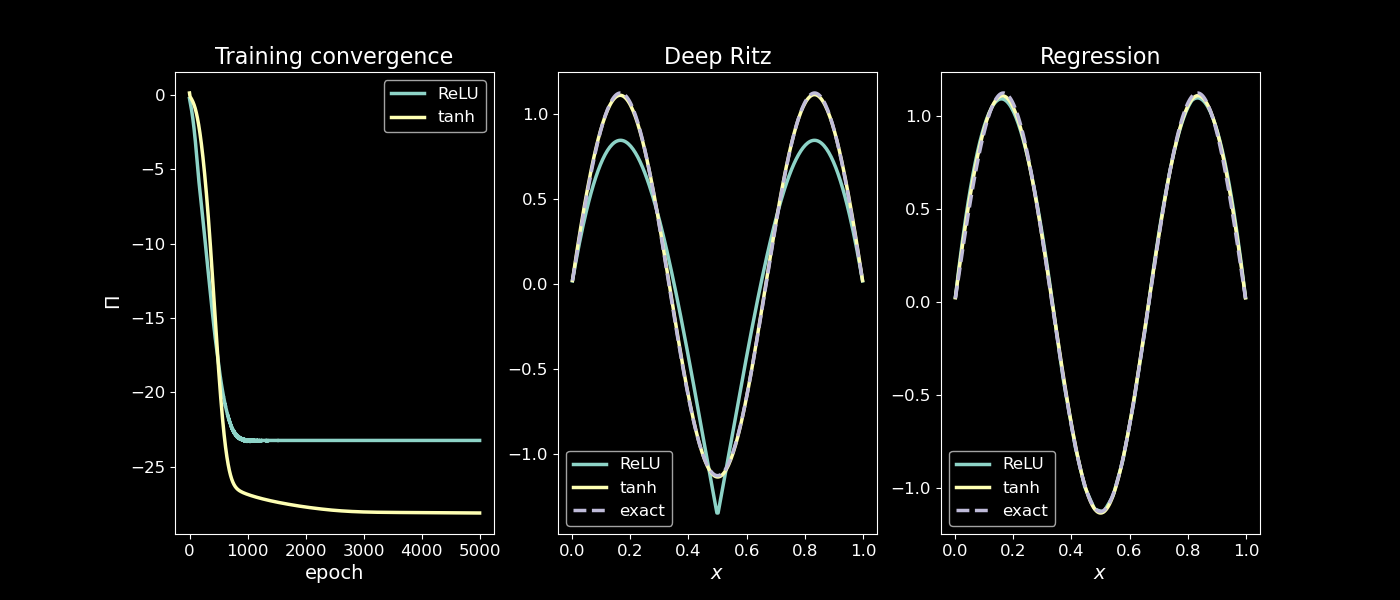}
\caption{Training convergence of the variational energy objective for the two networks (left). The Deep Ritz solution with ReLU activation fails to capture the true solution (center). However, when trained explicitly to represent the true solution by minimizing the mean squared error in a regression problem, the ReLU network performs well (right).}
\label{provocation}
\end{figure}

\begin{figure}[hbt!]
\centering
\includegraphics[width=0.99\textwidth]{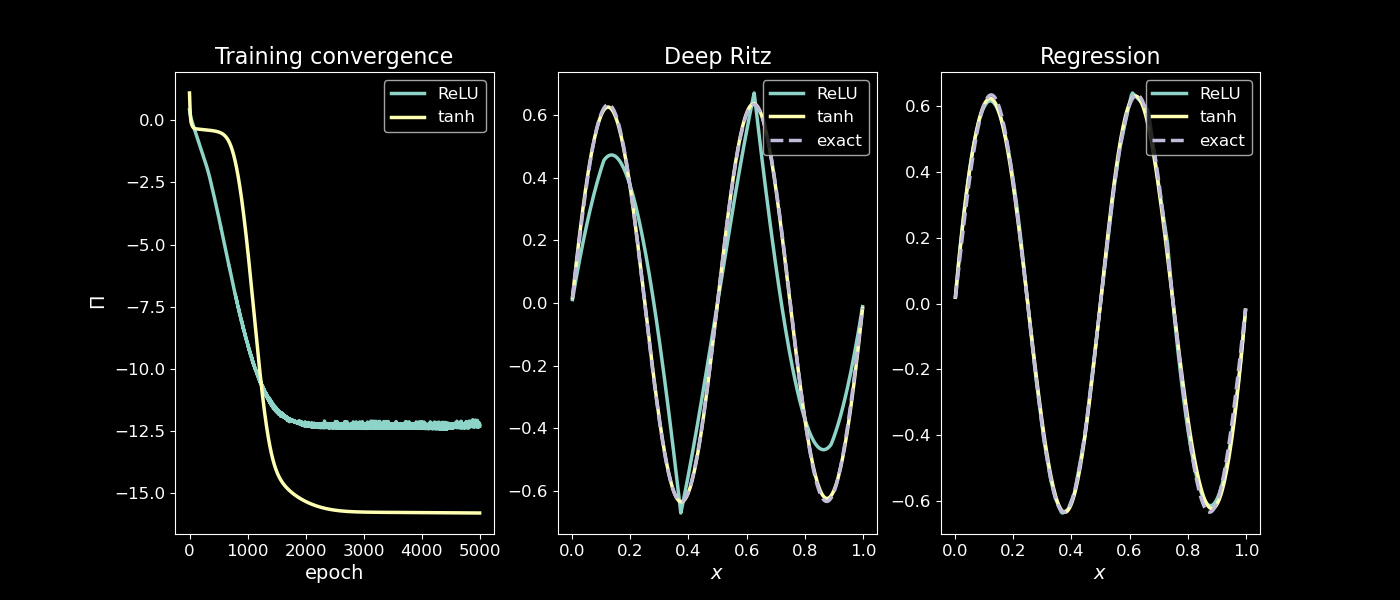}
\caption{Training convergence of the variational energy objective with a higher-frequency source term for the two networks (left). The Deep Ritz solution with ReLU activation again provides an unphysical and inaccurate representation of the solution (center). We verify that the network is sufficiently expressive to represent the desired solution field (right).}
\label{provocation2}
\end{figure}

\paragraph{} We use ADAM optimization with a learning rate of $1 \times 10^{-3}$ to minimize the variational energy. Integration is done on a uniform grid of $250$ points and we run the optimization for $5000$ epochs. The exact solution is obtained by inspection as $\frac{100}{9\pi^2} \sin( 3 \pi x)$. As an additional point of reference, we train a second single hidden-layer network of the same width with hyperbolic tangent activation functions. See Figure \ref{provocation} for the results. The converged energy of the ReLU network is notably larger than that of the network using hyperbolic tangent activation functions. The eyeball norm suffices to show that the ReLU network fails to reproduce the simple solution field. As the figure shows, this failure occurs despite the fact that the network can represent the exact solution when explicitly trained to do so. Figure \ref{provocation2} shows similar results for the higher-frequency source term $b(x) = 100 \sin( 4 \pi x)$. Evidently, these failures are not issues of expressiveness. We remark that the jagged shape of the solution field is similar to results obtained in \cite{dung_choice_nodate}, where ReLU provides very poor approximations of solution fields from incompressible flow problems. If the ReLU network is expressive enough to represent the true solution, and it possesses the same continuity properties as finite element discretizations, why does it perform so poorly with the Deep Ritz method? This is the question we address in the subsequent section.

\section{Diagnosing the failure of ReLU in the Deep Ritz problem}

\paragraph{} We continue to work with single hidden-layer networks for the sake of exposition, though the results we obtain are valid for deeper networks. Though the Deep Ritz objective requires only first derivatives of the solution, we also require parameter gradients of the network for training. The parameter gradient of the discretized Deep Ritz objective is 

\begin{equation}\label{gradient}
    \pd{\Pi}{\boldsymbol \theta} = \int_0^1 \kappa(x) \pd{u}{x} \frac{\partial^2 u}{\partial x \partial \boldsymbol \theta} - b(x) \pd{u}{\boldsymbol \theta} dx,
\end{equation}

\noindent where the discretized solution is given by Eq. \eqref{1hldist}. This gradient is used in the training process to iteratively update the parameters, but is often ``out of sight'' for the analyst, as it is computed by automatically differentiating the forward evaluation of the energy. We now compute each of the terms involved in Eq. \eqref{gradient} analytically. Recall that the solution field is given by the single hidden-layer network as 

\begin{equation*}
    u(x) = \sum_{i=1}^N \theta^3_i \sin(\pi x)R( \theta^1_i x + \theta^2_i)
\end{equation*}

The spatial gradient of the solution is thus

\begin{equation}\label{space}
    \pd{u}{x} = \sum_{i=1}^N \theta_i^3\Big( \pi \cos(\pi x) R(\theta^1_ix + \theta^2_i) + \sin(\pi x) \pd{R}{x}(\theta^1_i x + \theta_i^2) \theta^1_i \Big).
\end{equation}

The parameter gradients need to be broken out into gradients with respect to $\boldsymbol \theta^1$, $\boldsymbol \theta^2$, and $\boldsymbol \theta^3$ respectively. The gradient of the solution with respect to the parameters is 

\begin{equation}\label{parameter}
    \begin{aligned}
        \pd{u}{\boldsymbol \theta} = [ \partial u / \partial \boldsymbol \theta^1 , \partial u / \partial \boldsymbol \theta^2 , \partial u / \partial \boldsymbol \theta^3]^T,\\
        \pd{u}{\theta^1_j} = \theta^3_j \sin(\pi x) \pd{R}{x}(\theta^1_jx + \theta^2_j) x, \\
        \pd{u}{\theta^2_k} = \theta^3_k \sin(\pi x) \pd{R}{x}(\theta^1_k x + \theta^2_k) , \\
        \pd{u}{\theta^3_{\ell}} = \sin(\pi x) R(\theta^1_{\ell}x + \theta^2_{\ell} ).
    \end{aligned}
\end{equation}

Finally, the parameter gradient of the spatial gradient of the solution is 

\begin{equation}\label{mixed}
    \begin{aligned}
    \frac{\partial^2 u}{\partial x \partial \boldsymbol \theta} = [ \partial^2 u / \partial x \partial \boldsymbol \theta^1 , \partial^2 u / \partial x \partial \boldsymbol \theta^2 , \partial^2 u / \partial x\partial \boldsymbol \theta^3]^T,\\
    \frac{\partial^2 u}{\partial x \partial \theta^1_j} = \theta^3_j\Big( \pi \cos(\pi x) \pd{R}{x}(\theta^1_j x + \theta^2_j)x + \sin(\pi x) \theta^1_j \pdd{R}{x}(\theta^1_j x + \theta^2_j)x + \sin(\pi x) \pd{R}{x}(\theta^1_j x + \theta^2_j)
    \Big) , \\
    \frac{\partial^2 u}{\partial x \partial \theta^2_k} = \theta^3_k\Big( \pi \cos(\pi x) \pd{R}{x}(\theta^1_k x + \theta^2_k) + \sin(\pi x) \theta^1_k \pdd{R}{x}(\theta^1_k x + \theta^2_k)\Big) , \\
    \frac{\partial^2 u}{\partial x \partial \theta^3_{\ell}} = \pi \cos(\pi x) R( \theta^1_{\ell} x + \theta^2_{\ell} ) + \sin(\pi x) \theta^1_{\ell} \pd{R}{x}( \theta^1_{\ell} x + \theta^2_{\ell}).
    \end{aligned}
\end{equation}

The two derivatives of the ReLU activation are given by

\begin{equation}
    \pd{}{x}R(x) = \begin{cases}
        0, \quad x\leq 0, \\ 1, \quad x>0,
    \end{cases} \quad \pdd{}{x}R(x) = \delta(x).
\end{equation}

Notice that second derivatives of ReLU appear in Eq. \eqref{mixed}. This is a consequence of the nonlinearity of the discretization---while the $C^0$ continuity of the solution field is appropriate for a traditional Ritz method, training the network with the \textit{Deep} Ritz method requires second derivatives of the activation. This is because trainable parameters lie within the ReLU activation, unlike a so called ``linear'' discretization, where the parameters only multiply spatial basis functions. Mathematically speaking, the delta function arising from the second derivative of ReLU is acceptable given that it appears under the integral in the variational energy. However, it is not immediately clear how PyTorch's automatic differentiation handles gradients of discontinuous functions. We will investigate this numerically. To do this, we compute derivatives of the function $\sin(\pi x)\max(0,x-1/2)$ both analytically and with automatic differentiation. This function is reminiscent of one of the $N$ learned ``basis'' functions that are scaled and superimposed in the ReLU network of Eq. \eqref{1hldist}. To distinguish between the two cases of analytic and automatic differentiation, we call the analytically differentiated function $u(x)$ and the automatically differentiated function $w(x)$. Note that we can use Eqs. \eqref{space}-\eqref{mixed} with $N=1$, $\theta^1=1$, $\theta^2=-1/2$, and $\theta^3=1$ to obtain the relevant derivatives. To numerically approximate the delta function on a fixed integration grid, we require that

\begin{equation*}
    \int_0^1 \delta( x - \hat x) f(x) dx = f(\hat x) \rightarrow \Delta x\sum_{i=1}^{250} \delta_i f(x_i) \approx f( \hat x),
\end{equation*}

\noindent where $\delta_i$ is the discretized delta function and $\mathcal X = \{ x_i \}_{i=1}^{250}$ is the set of integration points with a corresponding fixed integration weight of $\Delta x$. This can be accomplished by writing the discretized delta as

\begin{equation}\label{discrete_delta}
    \delta_i= \begin{cases} 1/ \Delta x, \quad x_i = \text{argmin}(|\mathcal X - \hat x|), \\
    0, \quad \text{else}.
    \end{cases}
\end{equation}

\begin{figure}[hbt!]
\centering
\includegraphics[width=0.99\textwidth]{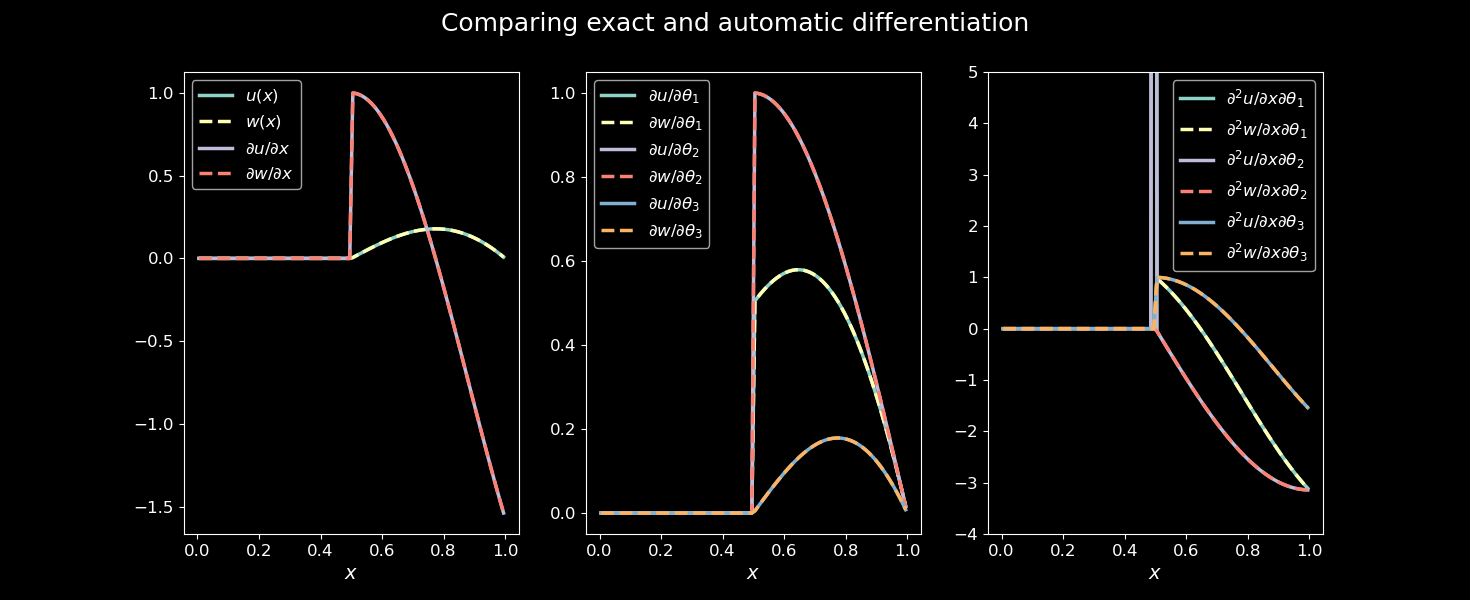}
\caption{Automatic differentiation handles the discontinuous spatial and parameter derivatives without issue, but, as shown in the case of mixed derivatives, the delta function arising from the derivative of a discontinuous function does not appear.}
\label{delta}
\end{figure}

\begin{figure}[hbt!]
\centering
\includegraphics[width=0.99\textwidth]{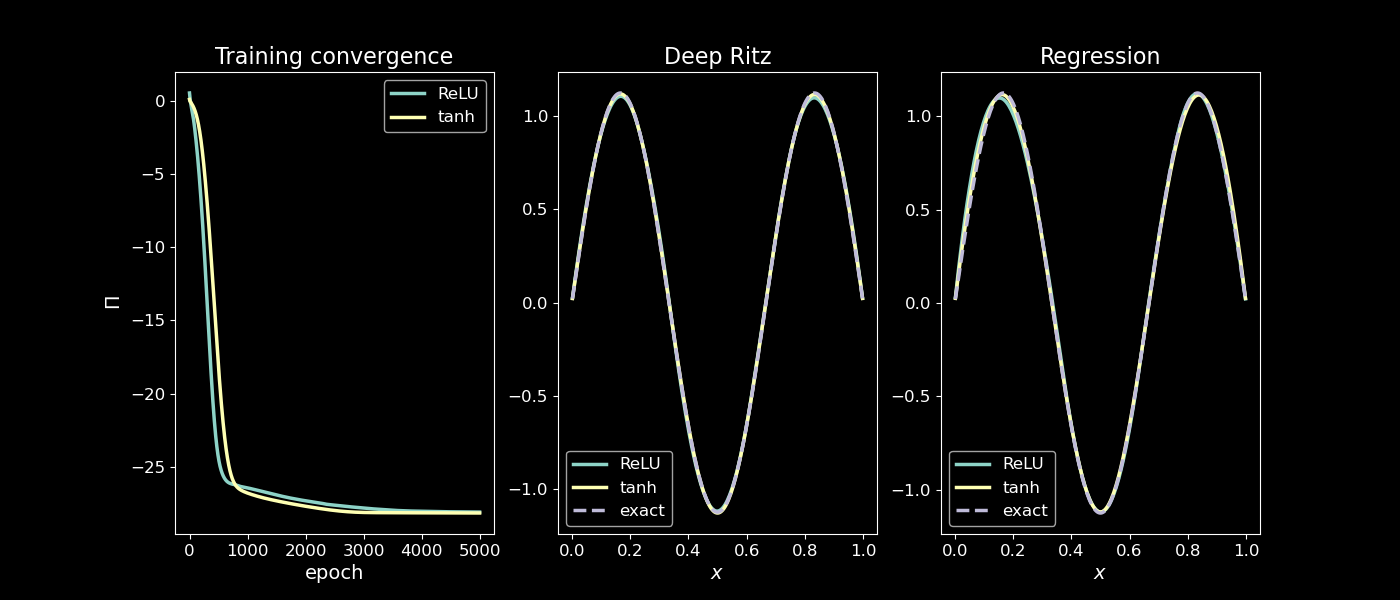}
\caption{Squaring the ReLU activation remedies the pathological solution obtained previously. This suggests that the $C^0$ continuity of ReLU is not sufficient for physics-informed training.}
\label{fixed}
\end{figure}

\begin{figure}[hbt!]
\centering
\includegraphics[width=0.99\textwidth]{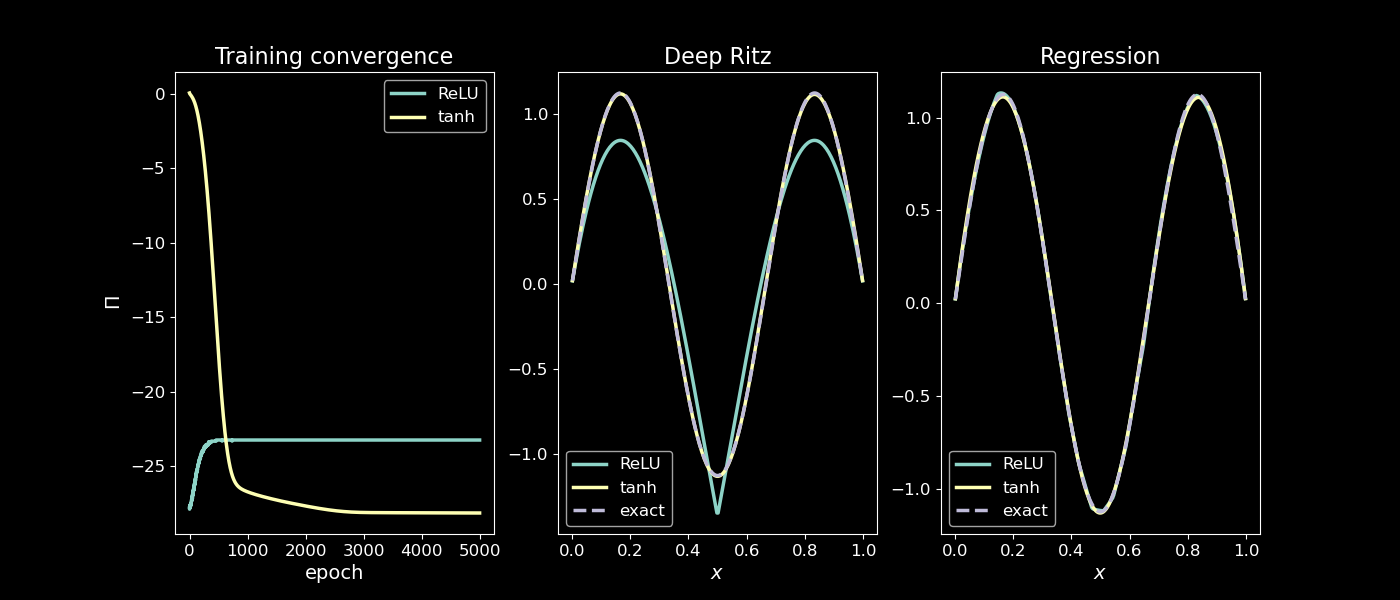}
\caption{When the standard ReLU network is initialized with the true solution field, the gradient updates increase the energy until the same inaccurate solution is obtained.}
\label{climbing}
\end{figure}

Eq. \eqref{discrete_delta} is how we numerically define the second derivative of ReLU centered at $\hat x$, which ensures that the discrete delta mimics the behavior of the continuous delta under numerical integration. Figure \ref{delta} shows the comparison of the spatial derivative, parameter gradients, and mixed spatial/parameter gradient obtained analytically and from automatic differentiation. Only in the mixed gradient do the second derivatives of ReLU appear, and we see that automatic differentiation neglects the delta function. Effectively, discontinuous functions are treated as $C^0$ continuous under automatic differentiation. This suggests that a possible explanation for the failure of ReLU networks in physics-informed problems is inaccurate gradient computations. It also suggests that introducing $C^1$ continuity would remedy the inaccuracy, as there would be no discontinuities to differentiate. We verify this hypothesis by solving the Deep Ritz problem with a ReLU-squared activation function given by $\max(0,x^2)$. Figure \ref{fixed} shows the results of re-running the original test problem with parameters given in Eq. \eqref{1hldist} and this updated activation. As expected, the ReLU-squared network performs equivalent to the network with hyperbolic tangent activation functions. Furthermore, we can show that initializing the ReLU network around the solution obtained from regression is not stationary under ADAM optimization carried out with automatic differentiation. This is shown in Figure \ref{climbing}. In fact, the inaccurate gradients push the ReLU network to the same higher-energy solution we obtained previously.

\begin{figure}[hbt!]
\centering
\includegraphics[width=0.99\textwidth]{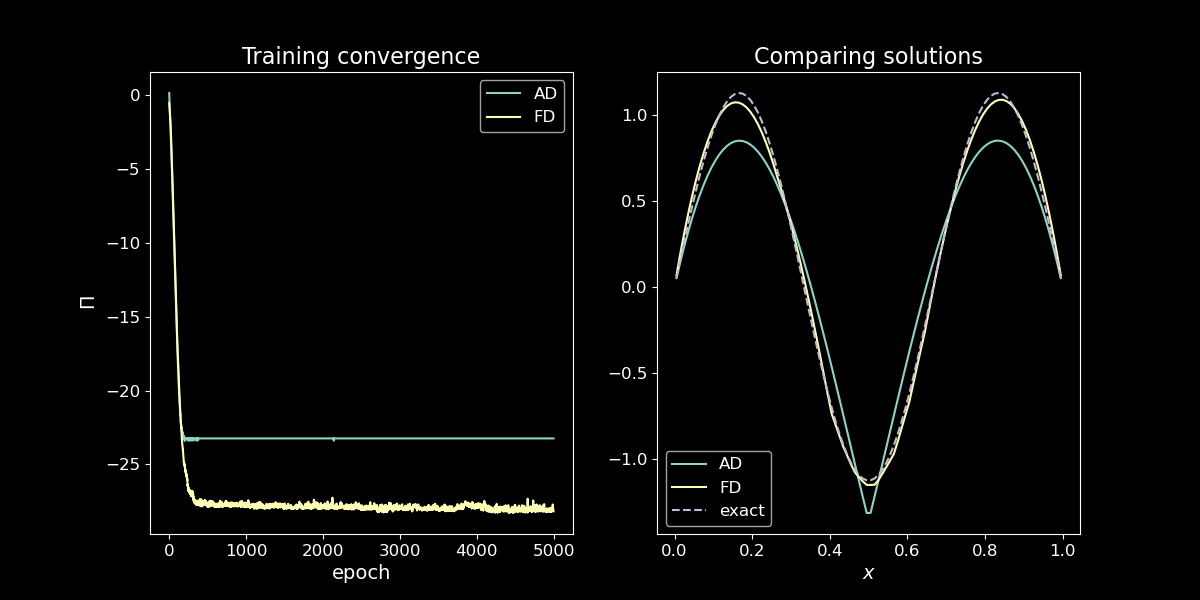}
\caption{The gradients computed using automatic differentiation cause early stagnation of the energy objective, whereas finite differenced gradients lead to a converged energy value which is comparable to the energy of the solution obtained from direct MSE fitting.}
\label{fd_ad}
\end{figure}

\paragraph{} As another verification of our hypothesis that automatic differentiation inaccurately computes gradients of the Deep Ritz loss function with ReLU activation, we train the network with gradients obtained from finite differencing. In this case, each component of the energy gradient is approximated using central differencing as

\begin{equation*}
    \pd{\Pi}{\theta_i} \approx \frac{1}{2h}( \Pi( \boldsymbol \theta+ h \mathbf{1}_i) - \Pi(\boldsymbol \theta - h \mathbf 1_i) ),
\end{equation*}

\noindent where the notation $\mathbf 1_i$ indicates a vector of zeros with a single $1$ in the $i$-th index. We take the finite difference step size to be $h=1 \times 10^{-3}$, the learning rate to be $5\times10^{-3}$, and run ADAM optimization for $5000$ epochs. The first single hidden-layer ReLU network is trained with gradients from automatic differentiation (AD) and the second is trained with manually supplied gradients computed using finite differences (FD). See Figure \ref{fd_ad} for the results of this comparison. The network with finite differenced gradients obtains a lower converged energy value, and a smoother, more accurate solution field.

\paragraph{} As a final inquiry into the use of ReLU in physics-informed contexts, we visualize the loss landscape defined by the variational energy of Eq. \eqref{energy}. We train a single hidden-layer ReLU network of width $50$ to represent the exact solution of the test problem. In particular, we define the optimal parameters as 

\begin{equation*}
    \boldsymbol{\tilde \theta} = \underset{\boldsymbol \theta}{\text{argmin }} \frac{1}{2} \int_0^1 \qty( \frac{100}{9 \pi^2} \sin( 3 \pi x) -  \sum_{i=1}^{50} \theta^3_i \sin(\pi x)R( \theta^1_i x + \theta^2_i) )^2 dx.
\end{equation*}

The Hessian matrix of the objective characterizes the curvature of the loss surface, and the eigenvectors of the Hessian are directions of extremal curvature. The Hessian matrix at the solution is 

\begin{equation*}
    \mathbf J  = \frac{\partial^2 \Pi}{\partial \boldsymbol \theta \partial \boldsymbol \theta}(\boldsymbol{\tilde \theta}).
\end{equation*}

\begin{figure}[hbt!]
\centering
\includegraphics[width=0.99\textwidth]{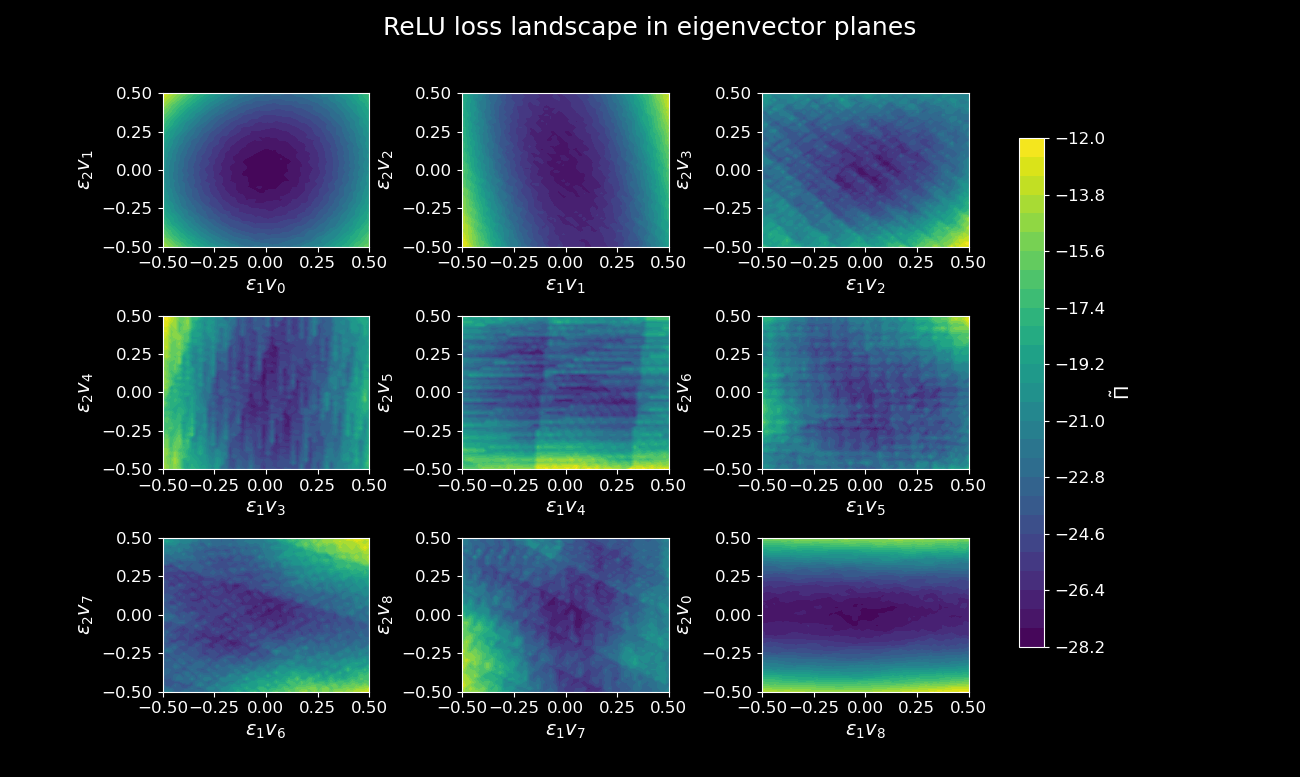}
\caption{The variational energy with ReLU activation around the true solution in planes spanned by eigenvectors of the Hessian. The lack of smoothness, and the consequent noise in gradients, is problematic considering that minima are found with iterative optimization strategies.}
\label{ReLU1}
\end{figure}

\begin{figure}[hbt!]
\centering
\includegraphics[width=0.99\textwidth]{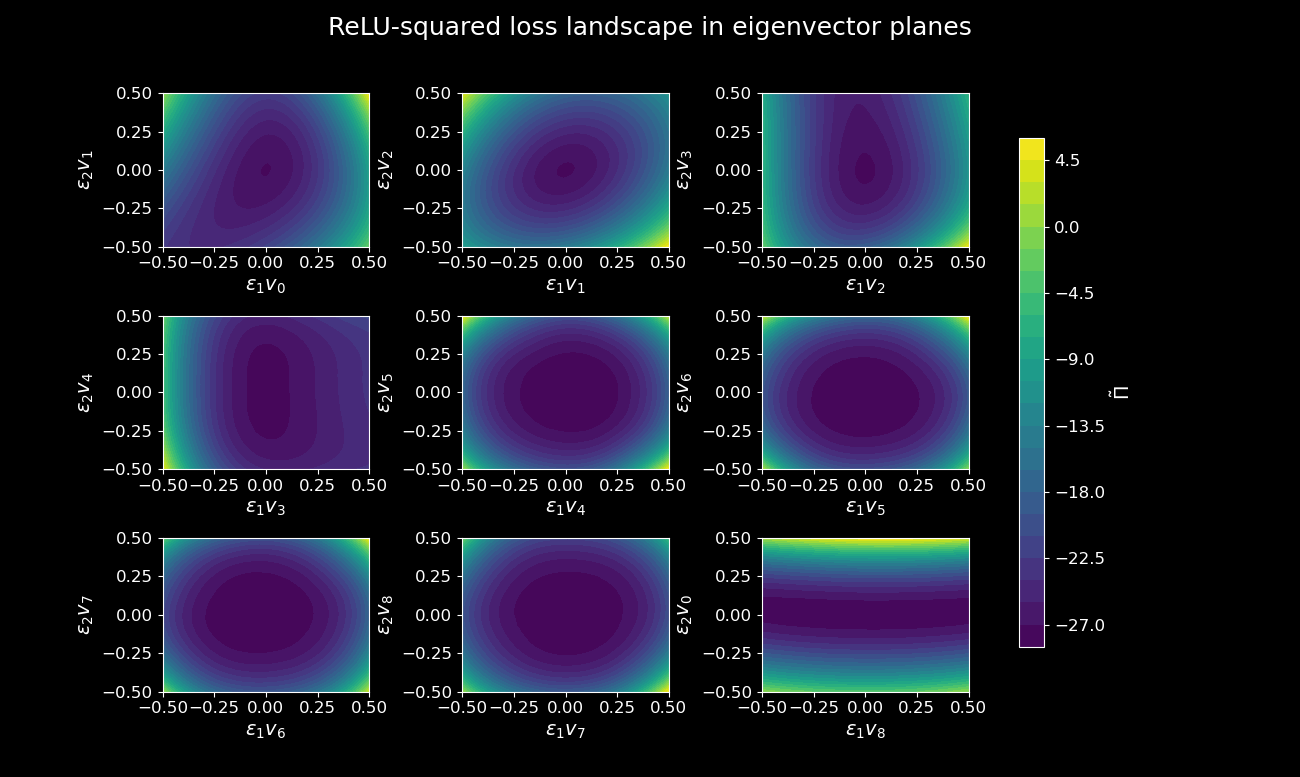}
\caption{When the ReLU activation is replaced with its square, the non-smooth features of the energy landscape disappear. We hypothesize this is a consequence of higher-order continuity.}
\label{ReLU2}
\end{figure}

This quantity can be computed using automatic differentiation, though the above discussion suggests this procedure may not be accurate. The potential inaccuracy is unimportant in this case, as we only wish to strategically choose cross-sections on which to visualize the loss. We call the eigenvectors of the Hessian $\mathbf v_1,\mathbf v_2,\dots$ which correspond to descending eigenvalues $\lambda_1,\lambda_2,\dots$. Per \cite{bottcher_visualizing_2024}, we visualize the loss landscape in planes spanned by eigenvectors of the Hessian. Specifically, we plot two-dimensional surfaces defined by

\begin{equation*}
    \tilde \Pi^{i,j}(\epsilon_1 , \epsilon_2) = \Pi( \boldsymbol{\tilde \theta} + \epsilon_1 \mathbf v_i + \epsilon_2 \mathbf v_j),
\end{equation*}

\noindent where $i$ and $j$ are indices of the eigenvectors and we take $\epsilon_1,\epsilon_2 \in [-1/2,1/2]$. Figure \ref{ReLU1} shows cross-sections of the energy landscape around the true solution obtained with nine pairs of eigenvectors. The loss is highly noisy and non-smooth. Contrast this with Figure \ref{ReLU2}, which is the same single hidden-layer network but with ReLU-squared activation. The higher-order continuity gives rise to a smoother loss landscape, indicating an easier optimization problem. Even if the issues with automatically differentiating discontinuous functions were remedied, the complex loss landscape arising from ReLU suggests that it is a poor choice for physics-informed problems.

\section{Conclusion}

\paragraph{} The purpose of this short work has been to explain failures of ReLU activation functions for physics-informed machine learning problems. Though the objective in the Deep Ritz test problem we considered required only first derivatives of the solution field, we showed that training the network requires second derivatives of the activation function. The second derivatives of the $C^0$ ReLU activation give rise to delta functions which automatic differentiation does not account for, thus leading to inaccurate gradient computations. We obtained accurate gradient computations by finite differencing the loss function, and showed that this remedied the failures of the ReLU network on the test problem. Though the corrected gradients improved performance, we then showed that the loss landscape of our test problem with the ReLU network is extremely noisy, and that the noise promptly vanishes when higher-order continuity is introduced. This study suggests that ReLU activation is a poor choice for physics-informed machine learning, which corroborates the conclusions of past comparisons of activation functions in the PINNs literature.



\begin{thebibliography}{10}

\bibitem{al-safwan_is_2021}
Ali Al-Safwan, Chao Song, and Umair~bin Waheed.
\newblock Is it time to swish? {Comparing} activation functions in solving the {Helmholtz} equation using physics-informed neural networks, October 2021.
\newblock arXiv:2110.07721 [physics].

\bibitem{bauchau_structural_2009}
O.~A. Bauchau and J.~I. Craig.
\newblock {\em Structural {Analysis}: {With} {Applications} to {Aerospace} {Structures}}.
\newblock Springer Science \& Business Media, August 2009.
\newblock Google-Books-ID: GYRX8ZYVNYQC.

\bibitem{bazant_critical_2022}
Zdeněk~P. Bažant, Hoang~T. Nguyen, and A.~Abdullah~Dönmez.
\newblock Critical {Comparison} of {Phase}-{Field}, {Peridynamics}, and {Crack} {Band} {Model} {M7} in {Light} of {Gap} {Test} and {Classical} {Fracture} {Tests}.
\newblock {\em Journal of Applied Mechanics}, 89(061008), April 2022.

\bibitem{brunton_discovering_2016}
Steven~L. Brunton, Joshua~L. Proctor, and J.~Nathan Kutz.
\newblock Discovering governing equations from data by sparse identification of nonlinear dynamical systems.
\newblock {\em Proceedings of the National Academy of Sciences}, 113(15):3932--3937, April 2016.
\newblock Publisher: Proceedings of the National Academy of Sciences.

\bibitem{bottcher_visualizing_2024}
Lucas Böttcher and Gregory Wheeler.
\newblock Visualizing high-dimensional loss landscapes with {Hessian} directions.
\newblock {\em Journal of Statistical Mechanics: Theory and Experiment}, 2024(2):023401, February 2024.
\newblock arXiv:2208.13219 [cs].

\bibitem{cranmer_interpretable_2023}
Miles Cranmer.
\newblock Interpretable {Machine} {Learning} for {Science} with {PySR} and {SymbolicRegression}.jl, May 2023.
\newblock arXiv:2305.01582 [astro-ph].

\bibitem{dung_choice_nodate}
Duong~V. Dung, Nguyen~D. Song, Pramudita~S. Palar, and Lavi~R. Zuhal.
\newblock On {The} {Choice} of {Activation} {Functions} in {Physics}-{Informed} {Neural} {Network} for {Solving} {Incompressible} {Fluid} {Flows}.
\newblock In {\em {AIAA} {SCITECH} 2023 {Forum}}. American Institute of Aeronautics and Astronautics.
\newblock \_eprint: https://arc.aiaa.org/doi/pdf/10.2514/6.2023-1803.

\bibitem{e_deep_2017}
Weinan E and Bing Yu.
\newblock The {Deep} {Ritz} method: {A} deep learning-based numerical algorithm for solving variational problems, September 2017.
\newblock arXiv:1710.00211 [cs].

\bibitem{fazliani_enhancing_2025}
Shaghayegh Fazliani, Zachary Frangella, and Madeleine Udell.
\newblock Enhancing {Physics}-{Informed} {Neural} {Networks} {Through} {Feature} {Engineering}, June 2025.
\newblock arXiv:2502.07209 [cs].

\bibitem{ghanem_stochastic_1991}
Roger~G. Ghanem and Pol~D. Spanos.
\newblock {\em Stochastic {Finite} {Elements}: {A} {Spectral} {Approach}}.
\newblock Springer, New York, NY, 1991.

\bibitem{goldstein_classical_2002}
Herbert Goldstein, Charles Poole, John Safko, and Stephen~R. Addison.
\newblock Classical {Mechanics}, 3rd ed.
\newblock {\em American Journal of Physics}, 70(7), July 2002.

\bibitem{hendrycks_gaussian_2023}
Dan Hendrycks and Kevin Gimpel.
\newblock Gaussian {Error} {Linear} {Units} ({GELUs}), June 2023.
\newblock arXiv:1606.08415 [cs].

\bibitem{horgan_end_2017}
John Horgan.
\newblock {\em The {End} {Of} {Science}}.
\newblock 2017.

\bibitem{hughes_thomas_j_r_finite_2000}
Thomas J.~R. Hughes.
\newblock The {Finite} {Element} {Method}, 2000.

\bibitem{jagtap_adaptive_2020}
Ameya~D. Jagtap, Kenji Kawaguchi, and George~Em Karniadakis.
\newblock Adaptive activation functions accelerate convergence in deep and physics-informed neural networks.
\newblock {\em Journal of Computational Physics}, 404:109136, March 2020.

\bibitem{lai_molecular_2018}
Rui Lai, Eric~D. Dodds, and Hui Li.
\newblock Molecular dynamics simulation of ion mobility in gases.
\newblock {\em The Journal of Chemical Physics}, 148(6):064109, February 2018.

\bibitem{maczuga_influence_2023}
Paweł Maczuga and Maciej Paszyński.
\newblock Influence of {Activation} {Functions} on the {Convergence} of {Physics}-{Informed} {Neural} {Networks} for {1D} {Wave} {Equation}.
\newblock In Jiří Mikyška, Clélia de~Mulatier, Maciej Paszynski, Valeria~V. Krzhizhanovskaya, Jack~J. Dongarra, and Peter~M.A. Sloot, editors, {\em Computational {Science} – {ICCS} 2023}, pages 74--88, Cham, 2023. Springer Nature Switzerland.

\bibitem{mavlanov_dynamic_2025}
Tulkin Mavlanov, Sherzod Khudainazarov, and Abdurakhmon Rayimov.
\newblock Dynamic simulation of a human body by frequency response data.
\newblock {\em AIP Conference Proceedings}, 3265(1):050016, April 2025.

\bibitem{raissi_physics-informed_2019}
M.~Raissi, P.~Perdikaris, and G.E. Karniadakis.
\newblock Physics-informed neural networks: {A} deep learning framework for solving forward and inverse problems involving nonlinear partial differential equations.
\newblock {\em Journal of Computational Physics}, 378:686--707, February 2019.

\bibitem{sirignano_dgm_2018}
Justin Sirignano and Konstantinos Spiliopoulos.
\newblock {DGM}: {A} deep learning algorithm for solving partial differential equations.
\newblock {\em Journal of Computational Physics}, 375:1339--1364, December 2018.
\newblock arXiv:1708.07469 [q-fin].

\bibitem{sitzmann_implicit_2020}
Vincent Sitzmann, Julien N.~P. Martel, Alexander~W. Bergman, David~B. Lindell, and Gordon Wetzstein.
\newblock Implicit {Neural} {Representations} with {Periodic} {Activation} {Functions}, June 2020.
\newblock arXiv:2006.09661 [cs].

\bibitem{so_higher-order-relu-kans_2024}
Chi~Chiu So and Siu~Pang Yung.
\newblock Higher-order-{ReLU}-{KANs} ({HRKANs}) for solving physics-informed neural networks ({PINNs}) more accurately, robustly and faster, September 2024.
\newblock arXiv:2409.14248 [cs].

\bibitem{sukumar_exact_2022}
N.~Sukumar and Ankit Srivastava.
\newblock Exact imposition of boundary conditions with distance functions in physics-informed deep neural networks.
\newblock {\em Computer Methods in Applied Mechanics and Engineering}, 389:114333, February 2022.

\bibitem{udrescu_ai_2020}
Silviu-Marian Udrescu and Max Tegmark.
\newblock {AI} {Feynman}: a {Physics}-{Inspired} {Method} for {Symbolic} {Regression}, April 2020.
\newblock arXiv:1905.11481 [physics].

\bibitem{wang_learning_2023}
Honghui Wang, Lu~Lu, Shiji Song, and Gao Huang.
\newblock Learning {Specialized} {Activation} {Functions} for {Physics}-informed {Neural} {Networks}.
\newblock {\em Communications in Computational Physics}, 34(4):869--906, June 2023.
\newblock arXiv:2308.04073 [cs].

\bibitem{westergaard_bearing_1939}
harold Westergaard.
\newblock Bearing pressure and cracks.
\newblock {\em Journal of applied mechanics}, 49, 1939.

\bibitem{zafar_optimizing_2025}
Husna Zafar, {Ahmad}, Xiangyang Liu, and Muhammad~Noveel Sadiq.
\newblock Optimizing {Physics}-{Informed} {Neural} {Networks} with hybrid activation functions: {A} comparative study on improving residual loss and accuracy using partial differential equations.
\newblock {\em Chaos, Solitons \& Fractals}, 191:115727, February 2025.

\end{thebibliography}

\end{document}